\pdfoutput=1
\documentclass[11pt]{article}
\usepackage{coling2020}
\usepackage{times}
\usepackage{url}
\usepackage{latexsym}

\usepackage{todonotes}
\usepackage{makecell}
\usepackage{changes}

\usepackage{booktabs}
\usepackage{array,graphicx}
\usepackage{booktabs}
\usepackage{pifont}
\usepackage{subfig}
\usepackage{color, colortbl}

\newcommand*\rot{\rotatebox{90}}
\newcommand*\CK{\ding{51}}

\colingfinalcopy 

\title{Neural Unsupervised Domain Adaptation in NLP---A Survey}

  \author{Alan Ramponi,{$^{1,2}$} Barbara Plank{$^{3}$} \\
  {$^{1}$}Department of Inf. Eng. and Computer Science, University of Trento, Italy \\
  {$^{2}$}Fondazione The Microsoft Research -- University of Trento \\Centre for Computational and Systems Biology (COSBI), Italy \\
  {$^{3}$}Department of Computer Science, ITU Copenhagen, Denmark \\
  {\tt ramponi@cosbi.eu, bplank@itu.dk} \\}

\date{}

\begin{document}
\maketitle
\begin{abstract}
    Deep neural networks excel at learning from labeled data and achieve state-of-the-art results on a wide array of Natural Language Processing tasks. In contrast, learning from unlabeled data, especially under domain shift, remains a challenge. 
    Motivated by the latest advances, in this survey we review neural unsupervised domain adaptation techniques which do not require labeled target domain data. This is a more challenging yet a more widely applicable setup. We outline methods, from early traditional non-neural methods to pre-trained model transfer. We also revisit the notion of \textit{domain}, and we uncover a bias in the type of Natural Language Processing tasks which received most attention. Lastly, we outline future directions, particularly the broader need for \textit{out-of-distribution generalization} of future NLP.\footnote{Accompanying repository: \url{https://github.com/bplank/awesome-neural-adaptation-in-NLP}}
\end{abstract} 

\section{Introduction}

Deep learning has undoubtedly pushed the frontier in Natural Language Processing (NLP). Particularly large pre-trained language  models have improved results for a wide range of NLP applications. However, the lack of portability of NLP models to new conditions remains a central issue in NLP. For many target applications, labeled data is lacking (Y scarcity), and even for pre-training general models data might be scarce (X scarcity). This makes it even more pressing to revisit a particular type of transfer learning, namely domain adaptation (DA).
A default assumption in many  machine learning algorithms is that the training and test sets follow the same underlying distribution. When these distributions do not match, we face a \textit{dataset shift}~\cite{gretton2007kernel} -- in NLP typically referred to as a \textit{domain shift}. In this setup, the \emph{target} domain and the \textit{source} training data differ, they are not sampled from the same underlying distribution. Consequently, performance drops on the target, which  undermines the ability of models to truly generalize \emph{into the wild}. Domain adaptation is closely tied to a fundamental bigger open issue in machine learning: generalization beyond the training distribution. Ultimately, intelligent systems should be able to adapt and robustly handle any test distribution, without having seen any data from it. This is the broader need for \textit{out-of-distribution generalization}~\cite{bengio-turinglec2019}, and a more challenging setup targeted at handling \textit{unknown} domains~\cite{volpi2018generalizing,krueger2020out}.

Work on domain adaptation focused largely on \textit{supervised} domain adaptation~\cite{daume2007easy,Plank2011}. In such a classic supervised DA setup, a small amount of labeled target domain data is available, along with some larger amount of labeled source domain data. The task is to adapt from the source to the specific target domain in light of limited target domain data. 
However, annotation is a substantial time-requiring and costly manual effort. While annotation directly mitigates the lack of labeled data, it does not  easily scale to new application targets.  In contrast, DA methods aim to shift the ability of models from the traditional interpolation of similar examples to models that extrapolate to examples outside the original training distribution \cite{Ruder2019}. \textit{Unsupervised domain adaptation} (UDA) mitigates the domain shift issue by learning only from \textit{unlabeled} target data, which is typically available for both source and target domain(s).
UDA fits the classical real-world scenario better, in which labeled data in the target domain is absent, but unlabeled data might be abundant.
UDA thus provides an elegant and scalable solution. We believe these advances in UDA will help for out-of-distribution generalization.

 \begin{figure}[!t]\centering
 	\includegraphics[width=\columnwidth]{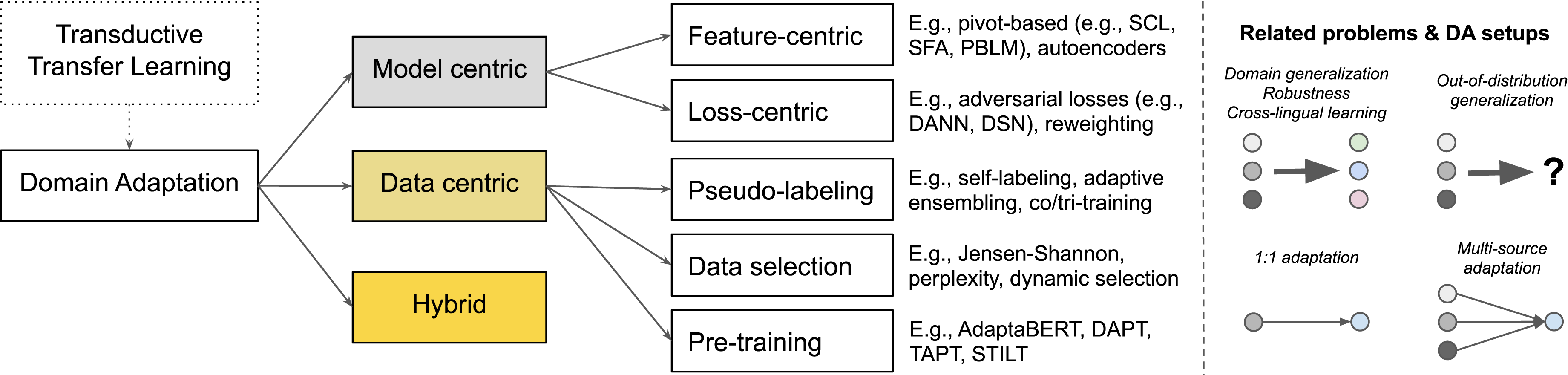}
 	\caption{Taxonomy of DA as special case of transductive transfer learning (left). Related problems (e.g., domain and out-of-distribution generalization) and DA setups (1:1 and multi-source adaptation) (right).}
 	\label{fig:taxonomy}
 \end{figure}
 
\paragraph{A categorization of domain adaptation in NLP} 
We categorize research into model-centric, data-centric and hybrid approaches, as shown in Figure~\ref{fig:taxonomy}.  \textit{Model-centric} methods target approaches to augment the feature space, alter the loss function, the architecture or model parameters~\cite{Blitzer2006,Pan2010,Ganin2016}. \textit{Data-centric} methods focus on the data aspect and either involve pseudo-labeling (or bootstrapping) to bridge the domain gap~\cite{Abney2007,Zhu:Goldberg:2009,ruder-plank-2018-strong,cui-bollegala-2019-self}, data selection~\cite{axelrod-etal-2011-domain,plank-van-noord-2011-effective,ruder-plank-2017-learning} and pre-training methods~\cite{Han2019,guo2020multisource}. As some approaches take elements of both, we include a \textit{hybrid} category.\footnote{We take inspiration of the data-centric and model-centric terms from~\newcite{Chu2018} in MT, and add hybrid.} A comprehensive overview of UDA methods and the tasks each method is applied to is provided in Table~\ref{tab:overview}.

\paragraph{Other surveys} Comprehensive reviews on DA exist, each with a different focus: visual applications \cite{Csurka2017,Patel2015,wilson_survey_2020}, machine translation (MT) \cite{Chu2018}, pre-neural DA methods in NLP \cite{jiang2008literature,Margolis2011}. Seminal surveys in machine learning on transfer learning include~\newcite{pan2009survey},~\newcite{weiss-ea-survey}, and~\newcite{yang_zhang_dai_pan_2020}.

\paragraph{Contributions} In this survey, we (i) comprehensively review neural approaches to unsupervised domain adaptation in NLP,\footnote{We disregard methods which are task-specific (like leveraging a sentiment thesaurus).} (ii) we analyze and compare the strengths and weaknesses of the described approaches, and (iii) we outline potential challenges and future directions in this field.

\section{Background} \label{sec:background}
First we introduce the classic learning paradigm with its core assumption, then we outline DA setups.
Given $\{ x_1, ..., x_n \} = X$ the training instances and $\{ y_1, ..., y_n \} = Y$ the corresponding class labels, the goal of machine learning is to learn a function $f$ that generalizes well to unseen instances. In \emph{supervised learning}, training data consists of tuples $\{ ( x_i, y_i ) \}_{i=1}^{n}$, where $n$ is the number of instances, while in \emph{unsupervised learning} we only have $\{ ( x_i ) \}_{i=1}^{n}$.
A general assumption in supervised machine learning is that the test data follows the same distribution as the training data. Formally, training and test data are assumed to be independently and identically (i.i.d.) sampled from the same underlying distribution. In practice, this assumption does not hold, which translates into a drop in performance when the model $f$ trained on a source domain $S$ is tested on a different but related target domain $T$.

\subsection{Domain adaptation and transfer learning: notation}

Formally, a domain is defined as $\mathcal{D} = \{ \mathcal{X}, P(X) \}$ where $\mathcal{X}$ is the feature space (e.g., the text representations), and $P(X)$ is the marginal probability distribution over that feature space. A task (e.g., sentiment classification) is defined as $\mathcal{T} = \{ \mathcal{Y}, P(Y|X) \}$, where $\mathcal{Y}$ is the label space. Estimates for the prior distribution $P(Y)$ and the likelihood $P(Y|X)$ are learned from the training data $\{ ( x_i, y_i ) \}_{i=1}^{n}$.

Domain adaptation aims to learn a function $f$ from a source domain $\mathcal{D}_S$ that generalizes well to a target domain $\mathcal{D}_T$, where $P_S(X) \neq P_T(X)$. DA is a particular case of transfer learning, namely \textit{transductive transfer learning}~\cite{pan2009survey,Ruder2019}. In inductive learning, the source and target tasks differ~\cite{pan2009survey}. In transductive DA, the source and target tasks $\mathcal{T}_S$ and $\mathcal{T}_T$ remain the same, 
but the source and target domains $\mathcal{D}_S$ and $\mathcal{D}_T$ 
differ in their underlying probability distributions. Given two distributions $P_S(X,Y)$ and $P_T(X,Y)$, DA typically addresses the shift in marginal distribution $P_S(X) \neq P_T(X)$, also known as covariate shift. A related problem is the problem of label shift, $P_S(Y) \neq P_T(Y)$. Since we do not assume any labeled target data, we focus on the former.\footnote{As outlined in~\newcite{Plank2011}, there exists a three-way distinction for domain adaptation: supervised DA, unsupervised DA but also \textit{semi-supervised} DA. The latter was coined around 2010 to distinguish purely unsupervised DA from cases where a small amount of labeled data is available in addition to the unlabeled target data. As this setup still assumes some labeled data and it has not received much attention, it is not discussed in the current survey.}

\section{What is a \textit{domain}? From the notion of domain to {variety space} and related problems}\label{sec:varietyspace}

Despite the formal definition of \textit{domain} above, the term is quiet loosely used in NLP and there is no common ground on what constitutes a domain~\cite{Plank2016}. Typically in NLP, domain is meant to refer to some coherent type of corpus, i.e., predetermined by the given dataset~\cite{Plank2011}. This may relate to topic, style, genre, or linguistic register.  The notion of domain and what plays into it has though significantly changed over the last years, leading to relevant research lines.

First, the Penn Treebank WSJ corpus~\cite{marcus-etal-1993-building} and the Brown corpus~\cite{francis79browncorpus} are prototypical examples, with the WSJ being considered widely as the canonical newswire domain. 
In the recent decade, there has been considerable work on what is considered \textit{non-canonical} data. The dichotomy between canonical (typically considered well-edited English newswire)  and non-canonical data arose with the increasing interest of working with \textit{social media} with all its challenges related to the `noisiness' of the domain~\cite{eisenstein2013bad,baldwin-etal-2013-noisy}.
Models trained on canonical data failed in light of the challenges on, e.g., Twitter~\cite{gimpel2011part,foster2011hardtoparse}. 

The general quest to understand the implications of variations of language on model performance led to  lines of work on how human factors impact data in a covert or overt way, e.g., on how latent socio-demographic factors impact NLP performance~\cite{hovy-2015-demographic,nguyen2016computational}, or how direct data collection strategies like crowdsourcing impact corpus composition~\cite{geva-etal-2019-modeling} or frequency effects impact NLP performance~\cite{zhang-etal-2019-paws}.
However, \textit{what is a domain?} Is, say, Twitter, its own domain? Or is it a set of subdomains? Similarly, do language samples of social groups (e.g., sociolects) form a domain or a set of subdomains? 

\paragraph{Variety space} We believe it is time to reconsider the notion of \textit{domain}, the use of the term itself, and raise even more awareness of the underlying variation in the data samples NLP works with. NLP is pervasively facing heterogeneity in data along many underlying (often unknown) dimensions.  
A theoretical notion put forward by~\newcite{Plank2016} is the \textit{variety space}.  In the variety space
a \textit{corpus} is seen as a subspace (subregion), a sample of the variety space. A corpus is a set of instances drawn from the underlying unknown high-dimensional variety space, whose dimensions (or latent factors) are fuzzy language and annotation aspects. These latent factors can be related to the notions discussed above,  such as genre (e.g., scientific, newswire, informal), 
sub-domain (e.g., finance, immunology, politics, environmental law, molecular biology) 
and socio-demographic aspects (e.g., gender), among other unknown factors, as well as stylistic or data sampling impacts (e.g., sentence length, annotator bias). 

In spirit of the variety space~\cite{Plank2016}, 
we suggest to use the more general term \textit{variety}, rather than domain, which pinpoints better to the underlying linguistic differences and their implications rather than the technical assumptions. Each corpus is inevitably biased towards a specialized language and some latent aspects. Understanding bias sources and effects, besides effects only~\cite{shah-etal-2020-predictive}, and documenting the known are the first important steps~\cite{bender2018data}, as is building broader, more varied corpora~\cite{ide2004american}. What we need more work on is to link the known to the unknown, and studying its impact. Doing so will ultimately help to not only overcome overfitting to overrepresented domains (e.g.,\ the  newswire bias~\cite{Plank2016}), but also work on robustness and ultimately out-of-distribution generalization, as described later on. 

Treating data as `just another input' to machine learning is very problematic. 
For example, it is less known that the well-known Penn Treebank consists of multiple genres~\cite{webber-2009-genre,plank-van-noord-2011-effective}, including reviews and some prose. It has almost universally been treated as prototypical news domain. Similarly, social media is typically considered only non-canonical data, but an analysis revealed the data to lie on a ``continuum of similarity''~\cite{baldwin-etal-2013-noisy}. This has implications on NLP performance.
As we have seen, there are a multitude of dimensions to consider in corpus composition and annotations, which are tied to the theoretical notion of a variety space. They challenge the true generalization capabilities of current models. 
\textit{What remains is to study what variety comprises, how covert and overt factors impact results, and take them into consideration in modeling and evaluation.}

\paragraph{Related problems} Following the idea of the \textit{variety space}, we discuss three related notions:  \textit{cross-lingual learning}, \textit{domain generalization/robustness}, and \textit{out-of-distribution generalization}.  

In \textit{cross-lingual learning} the feature space drastically changes, as alphabets, vocabularies and word order can be different. It can be seen as extreme adaptation scenario, for which parallel data may exist and can be used to build multilingual representations~\cite{ruder2019survey,artetxe-etal-2020-call}. Second, instead of adapting to a particular target, there is some work on \textit{domain generalization} aimed at building a single system  which is robust on several known target domains. One example is the SANCL shared task~\cite{petrov2012overview}, where participants were asked to build a single system that can robustly parse reviews, weblogs, answers, emails, newsgroups. In this setup, the DA problem boils down to finding a more robust system for given targets. It can be seen as optimizing for both in-domain and out-of-domain(s) accuracy.

If domains are unknown a priori, robustness can be taken a step further towards \textit{out-of-domain generalization}, to unknown targets, the most challenging setup. A recent solution is \textit{distributionally robust optimization}~\cite{oren-etal-2019-distributionally}, i.e., optimizing for worst-case performance without the knowledge of the test distribution. To do so, it assumes a \textit{subpopulation shift}, where the test population is a subpopulation mix of the training distribution. A model is then trained to do well over a wide range of potential test distributions. Some early work in dialogue~\cite{bod1999context} and parsing~\cite{plank2008subdomain} adopted a similar idea of \textit{subdomains}, however, with manually identified subpopulations.  This bears some similarity to early work on leveraging general background knowledge (embeddings trained on general data) for domain adaptation~\cite{Plank2013,Nguyen2015,Li2018}, and also relates to recent work on pre-training (Section~\ref{sec:pre-training}). An alternative and complementary interesting line of research is to \textit{predict test set performance} for new data varieties~\cite{ravi-etal-2008-automatic,van-asch-daelemans-2010-using,elsahar-galle-2019-annotate,xia2020predicting}.

\definecolor{modelc}{rgb}{0.86, 0.86, 0.86} 
\definecolor{datac}{rgb}{1.0, 0.84, 0.0} %
\definecolor{hybrid}{rgb}{0.93, 0.86, 0.51}
\begin{table}[ht!]
\centering
\small
\resizebox{1\linewidth}{!}{%
\begin{tabular}{|l|l|llll|llll|}
\toprule
            & & \multicolumn{4}{c|}{classif./inference} & \multicolumn{4}{c|}{struct.\ prediction}\\
Work              &   Method              & \rot{SA} & \rot{LI} & \rot{TC} &  \rot{NLI} & \rot{POS} & \rot{DEP} & \rot{NER} & \rot{RE} \\ 
\midrule
\rowcolor{modelc}
\textit{Model-centric:} &  &  & & & & & & &   \\
\rowcolor{modelc}

(Ziser and Reichart 2017; 2018a; 2018b; 2019) & Neural SCL & \CK  & & & & & & &   \\
\rowcolor{modelc}
\cite{Miller2019} & Neural SCL (Joint AE-SCL) & \CK  & & &  & & &    &\\
\rowcolor{modelc}
\cite{Glorot2011} & SDA  & \CK & & & & &  & &    \\
\rowcolor{modelc}
\cite{Chen2012} & MSDA  & \CK& & & & & &  &  \\
\rowcolor{modelc}
\cite{yang_fast_2014} & MSDA & &  & &  &  \CK &  &  &   \\ 
\rowcolor{modelc}
\cite{Clinchant2016} & MSDA & \CK  &  & \CK & & & &  &  \\
\rowcolor{modelc}

\rowcolor{modelc}
\cite{Ganin2016} & DANN & \CK & &  & & &  &  &\\
\rowcolor{modelc}
\cite{li_end--end_2017} & DANN+SCL MemNet & \CK &  & & & & & &     \\
\rowcolor{modelc}
\cite{kim_adversarial_2017}  & DANN/DSN & & & \CK &  &   & & \CK &   \\
\rowcolor{modelc}
\cite{sato-etal-2017-adversarial} & DANN & & & & &  & \CK &  &   \\  
\rowcolor{modelc}
\cite{wu-etal-2017-adversarial} & DANN & & & & &  &  &  &  \CK \\  
\rowcolor{modelc}
\cite{yasunaga-etal-2018-robust} & DANN & & & & & \CK &   &  &   \\  
\rowcolor{modelc}
\cite{Shen2018} & DANN & \CK& & & &  & &  & \\
\rowcolor{modelc}
\cite{Li2018} & DANN & \CK & \CK& & &  & &   & \\
\rowcolor{modelc}
\cite{Alam2018} & DANN & & & \CK & & &  &   & \\
\rowcolor{modelc}
\cite{wang_adversarial_2019} & DANN & & & \CK & & & & &  \\ 
\rowcolor{modelc}
\cite{Shah2018} & DANN+Wasserstein & & &  \CK &  & & & &  \\ 
\rowcolor{modelc}
\cite{Fu2017} & DANN & & & & & & & &  \CK  \\ 
\rowcolor{modelc}
\cite{Rios2018} & DANN & &  & & &  & &  &\CK  \\
\rowcolor{modelc}
\cite{xu_adversarial_2019} & DANN & & &\CK &   &  & & &\\ 
\rowcolor{modelc}
\cite{Shi2018} & DSN (GSN) & & &  & & &  &  & \CK  \\ 
\rowcolor{modelc}
\cite{rocha-lopes-cardoso-2019-comparative}* & DANN, Shared encoders & \CK & &  &  \CK & & &  &  \\
\rowcolor{modelc}
\cite{ghosal-etal-2020-kingdom} & DANN (concept embeddings) & \CK & &  & & &  & &  \\
\rowcolor{modelc}
\cite{naik-rose-2020-towards} & DANN (context embeddings) & & &  & & &  & \CK &  \\

\rowcolor{datac}
\textit{Data-centric:} &  &  & & & & & & &   \\
 \rowcolor{datac}
\cite{ruder-plank-2018-strong} & SSL, Multitask tri-training  & \CK & & & & \CK & & & \\
 \rowcolor{datac}
\cite{lim_aaai2020} & SSL   &  & & & & \CK & & & \\
 \rowcolor{datac}
\cite{rotman-reichart2019} & Deep self-training & & & & &  & \CK &  &   \\ 

\rowcolor{datac}
\cite{Han2019} & AdaptaBERT$\diamond$ &  & & & & \CK &  & \CK &  \\
\rowcolor{modelc}
\rowcolor{datac}
\cite{li_semi-supervised_2019} & Adaptive pre-training & & & & &  & \CK &  &   \\
\rowcolor{datac}
\cite{gururangan2020dont} & Adaptive pre-training (incl.\ multi-phase)& \CK  & & \CK & \CK & & &  &  \CK \\
\rowcolor{hybrid}
\textit{Hybrid:} &  &  & & & & & & &   \\
\rowcolor{hybrid}
\cite{saito2017asymmetric} & Asymmetric tri-training &  \CK & & & &  & & & \\
\rowcolor{hybrid}
\cite{desai-etal-2019-adaptive} & Adaptive (temporal) ensembling & & & \CK &   & & &  &  \\
\rowcolor{hybrid}
\cite{jia-etal-2019-cross} & Cross-domain LM &  & & & & & & \CK &  \\
\rowcolor{hybrid}
\cite{cui-bollegala-2019-self} & SelfAdapt (pivots+co-training) & \CK & & & & & &  &   \\
\rowcolor{hybrid}
\cite{Peng2017} & Multi-task-DA$\diamond$ &  &  & & & \CK & & \CK &  \\ 
\rowcolor{hybrid}
\cite{guo2020multisource} & DistanceNet-Bandit & \CK & & & & & &  &   \\
\rowcolor{hybrid}
\cite{bendavid2020} & PERL (pivots+context embeddings) & \CK & & & & & &  &   \\
\bottomrule
\end{tabular}
}%
\caption{Overview of neural UDA in NLP: method and task(s).     Methods: SCL = structural correspondence learning; AE = autoencoder; SDA = stacked denoising autoencoder; MSDA = marginalized SDA; DANN = domain-adversarial neural network; DSN = domain separation network; GSN = genre separation network; SSL = semi-supervised learning; LM = language modeling. Tasks: SA = sentiment analysis; LI = language identification; TC = binary text classification (incl.\ machine reading, duplicate question detection, stance detection, intent classification, political data identification); NLI = natural language inference; POS = part-of-speech (incl.\ Chinese word segmentation); DEP = dependency parsing; NER = named entity recognition (incl.\ slot tagging, event trigger identification, named entity segmentation); RE = relation extraction. *with \emph{cross-lingual} adaptation. $\diamond$ applicable to UDA but main focus is supervised DA.}
\label{tab:overview}

\end{table}

\section{Model-centric approaches} \label{sec:model-centric}

Model-centric approaches redesign parts of the model: the feature space, the loss function or regularization and the structure of the model. We categorize them into feature-centric and loss-centric methods.

\subsection{Feature-centric methods} \label{sec:feature-centric}

Two lines of work can be found within feature-centric methods: \textit{feature augmentation} and \textit{feature generalization} methods. The former use \textit{pivots} (common shared features) to construct an aligned feature space. The latter use \textit{autoencoders} to find latent representations that transfer better across domains.

\paragraph{Pivots-based DA} 
Seminal \textit{pivot-based} methods include: \textit{structural correspondence learning} (SCL) \cite{Blitzer2006} and \textit{spectral feature alignment} (SFA) \cite{Pan2010}. They both aim at finding features which are common across domains by using unlabeled data from both domains. The two approaches differ in the specifics of the method to construct the shared space. SCL uses auxiliary functions inspired by~\newcite{ando2005framework}, while SFA uses a graph-based spectral learning method. Creating domain-specific and domain-general features is the key idea of EasyAdapt~\cite{daume2007easy}, a seminal supervised DA method.
A recent line of work~\cite{Ziser2017,Ziser2018,Ziser2018a,Ziser2019} brings SCL back to neural networks. 

In particular, \newcite{Ziser2017} propose to combine the strengths of pivot-based methods with autoencoder neural networks in an \textit{autoencoder structural correspondence learning} (AE-SCL) model. Autoencoders are used to learn latent representations to map non-pivots to pivots, and these encodings are then used to augment the training data. 
The main drawback of this approach is that the output vector representations of the text are unique and not context-dependent. To solve this problem, a \textit{pivot-based language modeling} (PBLM) method has been proposed \cite{Ziser2018,Ziser2018a}. PBLM effectively combines SCL with a neural language model based on long short-term memory (LSTM) networks which predicts the presence of pivots and non-pivots, thus making representations structure-aware. A weakness of the PBLM approach relies in the large number of pivots needed. To remedy this issue, Ziser and Reichart~\shortcite{Ziser2019} adopted a \textit{task refinement learning} approach using PBLM (called TRL-PBLM), showing gains in both accuracy and stability over different hyperparameters selection choices. The approach is an iterative training process where the network is trained using an increasingly larger amount of pivots. Recent hybrid UDA work extends pivots with contextual embeddings~\cite{bendavid2020}, as we discuss in Section~\ref{sec:hybrid}.

A common issue with the aforementioned methods is that they involve two independent steps: one for representation learning and one for task learning. To tackle this issue, recent studies propose training the two tasks jointly (i.e., pivot prediction and sentiment)~\cite{Miller2019} and learn pivots \textit{automatically} via attention~\cite{li_end--end_2017}, similar to work on automatic non-pivot identification~\cite{li2018nonpivots}. 

To the best of our knowledge, neural pivot-based UDA approaches have been solely applied to sentiment classification, cf.\ Table~\ref{tab:overview}. Notably, Ziser and Reichart~\shortcite{Ziser2018} went a step further, and applied neural SCL cross-lingually; the NLP task is still sentiment classification. The effectiveness of pivot-based methods in neural models remains to be tested. 
Early non-neural work applied SCL to structure prediction problems with mixed results, i.e., POS~\cite{Blitzer2006} and parsing~\cite{Plank2011}. 

\paragraph{Autoencoder-based DA} Early neural approaches for UDA have been based on autoencoders. Autoencoders are neural networks that are employed to learn latent representations from raw data in an unsupervised fashion by learning with an input reconstruction loss. Motivated by the \textit{denoising autoencoders} \cite{Vincent2008}, the first work in this line is by Glorot et al.~\shortcite{Glorot2011}, who introduced the \textit{stacked denoising autoencoder} (SDA) for domain adaptation. Basically, a SDA automatically learns a robust and unified feature representation for all domains by stacking multiple layers, and artificially corrupts the inputs with a Gaussian noise that the decoder needs to reconstruct. However, SDAs showed issues in speed and scalability to high-dimensional data. To mitigate these limitations, a more efficient \textit{marginalized stacked denoising autoencoder} (MSDA) that marginalizes the noise was proposed \cite{Chen2012}. MSDAs have been further extended by Yang and Eisenstein~\shortcite{yang_fast_2014} with marginalized structured dropout, and by Clinchant et al.~\shortcite{Clinchant2016}, which improved the regularization of MSDAs following the insights from the domain adversarial training of neural networks \cite{Ganin2015,Ganin2016} (described in Section \ref{sec:adversarial}). The main drawback of autoencoder approaches is that the induced representations do not make use of any linguistic information.

\subsection{Loss-centric methods}\label{sec:loss-centric}

Loss-centric approaches can be divided into methods which employ domain adversaries, and instance-level reweighting methods. We outline these two strands of work in the following.

\paragraph{Domain adversaries} \label{sec:adversarial}

The most widespread methods for neural UDA are based on the use of \textit{domain adversaries} \cite{Ganin2015,Ganin2016}. Inspired by the way generative adversarial networks (GANs) \cite{Goodfellow2014} minimize the discrepancies between training and synthetic data distributions, domain adversarial training aims at learning latent feature representations that serve at reducing the discrepancy between the source and target distributions. The intuition behind these methods puts its ground on the theory on domain adaptation \cite{Ben-David2010}, which argues that cross-domain generalization can be achieved by means of feature representations for which the origin (domain) of the input example cannot be identified.

The seminal approach in this category are DANNs: \textit{domain-adversarial neural networks}~\cite{Ganin2015,Ganin2016}. The aim is to estimate an accurate predictor for the task  while maximizing the confusion of an auxiliary domain classifier in distinguishing features from the source or the target domain. To learn domain-invariant feature representations, DANNs employ a loss function via a \textit{gradient reversal layer} which ensures that feature distributions in the source and target domains are made similar. The strength of this approach is in its scalability and generality; 
however, DANNs only model feature representations that are shared across both domains, and suffer from a vanishing gradient problem 
when the domain classifier accurately discriminates source and target representations \cite{Shen2018}. \textit{Wasserstein} methods~\cite{martin2017wasserstein} are more stable training methods than gradient reversal layers. Instead of learning a classifier to distinguish domains, they attempt to reduce the approximated Wasserstein distance (also
known as Earth Mover’s Distance). A recent study on question pair classification shows that the two adversarial methods reach similar performance, but Wasserstein enables more stable training~\cite{Shah2018}.

DANNs have been applied in many NLP tasks in the last few years, mainly to sentiment classification (e.g., Ganin et al.~\shortcite{Ganin2016}, Li et al.~\shortcite{Li2018}, Shen et al.~\shortcite{Shen2018}, Rocha and Lopes Cardoso~\shortcite{rocha-lopes-cardoso-2019-comparative}, Ghoshal et al.~\shortcite{ghosal-etal-2020-kingdom}, to name a few), 
but recently to many other tasks as well: language identification \cite{Li2018}, natural language inference~\cite{rocha-lopes-cardoso-2019-comparative}, POS tagging~\cite{yasunaga-etal-2018-robust}, parsing~\cite{sato-etal-2017-adversarial}, trigger identification~\cite{naik-rose-2020-towards}, relation extraction \cite{wu-etal-2017-adversarial,Fu2017,Rios2018}, and other (binary) text classification tasks like relevancy identification \cite{Alam2018}, machine reading comprehension \cite{wang_adversarial_2019}, stance detection~\cite{xu_adversarial_2019}, and duplicate question detection \cite{Shah2018}. This makes DANNs the most widely used UDA approach in NLP, as illustrated in Table~\ref{tab:overview}.

To model features that also belong to either the source or target domain, \textit{domain separation networks} (DSNs) \cite{Bousmalis2016} have been proposed. 
DSNs separate latent representations in i) separate private encoders (i.e., one for each domain) and ii) a shared encoder (in charge to reconstruct the input instance using these representations). This bears similarities to a traditional supervised  method~\cite{daume2007easy}. 
The main drawback of DSNs is that domain-specific representations are solely used in the decoder, leaving the classifier to be trained on the domain-invariant representations only.


DSNs have seen a notable success in Computer Vision (CV) \cite{Bousmalis2016}. In NLP, Shi et al.~\shortcite{Shi2018} propose the \textit{genre separation networks} (GSNs) as a variant of the DSNs, introducing a novel reconstruction component that leverages both shared and private feature representations in the learning process. As noted also by~\newcite{Han2019}, a downside of adversarial methods is that they require careful balancing between objectives~\cite{kim_adversarial_2017,Alam2018} to avoid instability during learning~\cite{pmlr-v70-arjovsky17a}.

\paragraph{Reweighting} \label{sec:reweighting} This family of methods is an instance-level adaptation method. The core idea of \textit{instance weighting} (also known as importance weighting) is to assign a weight to each training instance proportional to its similarity to the target domain~\cite{Jiang2007}. We can see instance weighting as an alternative to domain adversaries. While domain adversaries distinguish the domains to learn domain invariant representations in a joint model, instance weighting decouples domain detection for a-priori weight estimation of an instance.

Methods that explicitly reweight the loss based on domain discrepancy information include \textit{maximum mean discrepancy} (MMD)~\cite{gretton2007kernel} and its more efficient version called \textit{kernel mean matching} (KMM) \cite{gretton2009covariate}. KMM reweights the training instances such that the means of the training and test points in reproducing a kernel Hilbert
space are close to each other. \newcite{Jiang2007} introduced instance weighting in NLP and proposed to learn weights by first training domain classifiers. 
The effectiveness of the method in neural setups remains to be seen. An early study reports non-significant improvements for POS tagging~\cite{plank_importance_2014}.

\section{Data-centric methods}\label{sec:data-centric}

Recently, data-centric approaches are on a rise, due to rapid growth of data and the gain in popularity of pre-training methods. We summarize data-centric strands next, which differ whether they use pseudo-labeling, select relevant data or use large unlabeled data or auxiliary tasks for model pre-training. 

\subsection{Pseudo-labeling} The main idea of pseudo-labeling is to apply a trained classifier to predict labels on unlabeled instances, which are then treated as `pseudo' gold labels. Pseudo-labeling applies semi-supervised methods~\cite{Abney2007,Zhu:Goldberg:2009} such as bootstrapping methods like self-training, co-training and tri-training or  methods such as temporal ensembling~\cite{charniak1997statistical,McClosky2006,blum1998combining,Steedman2003,zhou2005tri,sogaard2010semi,saito2017asymmetric,laine2016temporal} by using either the same model, a teacher model, or multiple bootstrap models which may include slower but more accurate hand-crafted models~\cite{petrov-etal-2010-uptraining} to guide  pseudo-labeling. Most pseudo-labeling works date back to traditional non-neural learning methods. Bootstrapping methods for domain adaptation are well-studied in parsing~\cite{McClosky2006,reichart2007self,yu-etal-2015-domain}. They include models trained on other grammar formalisms to improve dependency parsing on Twitter~\cite{foster2011hardtoparse}. Recently, this line of classics has been revisited~\cite{ruder-plank-2018-strong,rotman-reichart2019,lim_aaai2020}. For example, classic methods such as tri-training constitute a strong baseline for domain shift in neural times~\cite{ruder-plank-2018-strong}.  Pseudo-labeling has recently been studied for parsing with contextualized word representations~\cite{rotman-reichart2019,lim_aaai2020} and a recent work  proposes \textit{adaptive ensembling}~\cite{desai-etal-2019-adaptive} as extension of temporal ensembling (see \textit{hybrid} methods in Section~\ref{sec:hybrid}).

\subsection{Data selection}\label{sec:dataselection}
A relatively unexplored area is data selection for adaptation, which is gaining traction again in light of large pre-trained models (which data should they be trained on?) and the related problem of cross-lingual learning (what is/are the best source language(s) to transfer from?).
Data selection aims to select the best matching data for a new domain, typically by using perplexity~\cite{moore-lewis-2010-intelligent} or using domain similarity measures such as Jensen-Shannon divergence over term or topic distributions~\cite{plank-van-noord-2011-effective}. This has mostly been studied for MT~\cite{moore-lewis-2010-intelligent,axelrod-etal-2011-domain,van-der-wees-etal-2017-dynamic,aharoni2020unsupervised}, but also for parsing~\cite{plank-van-noord-2011-effective,ruder-plank-2017-learning} and sentiment analysis~\cite{remus2012} though for supervised domain adaptation setups only. For parsing and sentiment analysis, the simple Jensen-Shannon divergence on term distribution constitutes a strong baseline~\cite{Plank2011,ruder-plank-2017-learning}. Within MT,~\newcite{van-der-wees-etal-2017-dynamic} propose a dynamic data selection approach which changes the subset of data in each epoch for MT.  Data selection is gaining attention, in light of the abundance of data. Recent work investigates data representation and cosine similarity for MT data selection~\cite{aharoni2020unsupervised}. Similarly, distance metrics have been been recently used for multi-source domain adaptation of sentiment classification models using a bandit-based approach~\cite{guo2020multisource}. For morphosyntactic cross-lingual work, simple overlap metrics are indicative~\cite{ustun-etal-2019-multi,lin-etal-2019-choosing}. Another line explores
whether tailoring large pre-trained models to the domain of a target task is still beneficial, and use of data selection to overcome costly expert selection. They propose two multi-phase pre-training methods~\cite{gururangan2020dont} (as discussed further below) with promising results on text classification tasks.

\subsection{Pre-training---And:---Is bigger better? Are domains (or: \textit{varieties}) still relevant?}\label{sec:pre-training}

Large pre-trained models have become ubiquitous in NLP~\cite{howard_universal_2018,peters-etal-2018-deep,devlin-etal-2019-bert}. \textit{Fine-tuning} a transformer-based model with a small amount of labeled data often reaches high performance across NLP tasks and has become a de-facto standard. It means starting from the pre-trained model weights and training a new task-specific layer on supervised data. A natural question which arises is how universal such large models are.
Is bigger better? And are domains (or varieties) still relevant? We return to these questions after depicting pre-training strategies. We delineate:

\begin{enumerate}
\itemsep0em 
    \item \textbf{Pre-training}: pre-training alone (e.g., multilingual BERT; language-specfic BERTs from scratch);
    \item \textbf{Adaptive pre-training}: This encompasses pre-training, followed by secondary stages of pre-training on unlabeled data or on labeled data from intermediate higher-resource auxiliary tasks:
    \begin{enumerate}
    \item \textbf{Multi-phase pre-training}:  two or more phases of secondary pre-training, from broad-coverage to domain-/task-adaptive pre-training (i.e., BioBERT, AdaptaBERT, DAPT, TAPT). They differ by the source of unlabeled data: broad-domain $\succ$ domain-specific $\succ$ task-specific;
    \item \textbf{Auxiliary-task pre-training}: pre-training, followed by (possibly multiple stages of) \textit{auxiliary-task} pre-training (e.g., supplementary training on intermediate labeled-data tasks, STILTs).
    \end{enumerate}
\end{enumerate}

\textit{Pre-training} (option 1) can be seen as straightforward adaptation, analogous to zero-shot in cross-lingual learning. The key idea is to train encoders  with self-supervised objectives like (masked) language model  and  related unsupervised objectives~\cite{peters-etal-2018-deep,devlin-etal-2019-bert,beltagy-etal-2019-scibert}. 

In light of a domain shift, \textit{adaptive pre-training} is beneficial, in which in one instantiation contextualized embeddings are adapted to text from the target domain by masked language modeling, as introduced by~\newcite{Han2019}.  
More broadly, we distinguish two variants of \textit{adaptive pre-training}. They differ whether unlabeled data or some form of auxiliary labeled data (or intermediate tasks data) is used. These variants can also be combined, and fine-tuning applies to all setups, if data is available. The key idea of \textit{multi-phase pre-training} (option 2a) is to use secondary-stage unsupervised pre-training, such as broad-coverage domain-specific BERT variants (e.g., BioBERT). ~\newcite{gururangan2020dont}  propose  \textit{domain-adaptive pre-training} (DAPT) from a broader corpus, compared to~\cite{Han2019}, and \textit{task-specific pre-training} (TAPT) which uses unlabeled data closer-and-closer to the task distribution. As these studies show, domain-relevant data is important for pre-training~\cite{Han2019,gururangan2020dont} in both high and low resource setups. Similar adaptive pre-training work has been shown to be effective for dependency parsing~\cite{li_semi-supervised_2019}. 
This suggests that there exists a spectrum of domains of varying granularity, confirming ideas around domain similarity~\cite{Plank2011,baldwin-etal-2013-noisy}. Domains (\textit{varieties}) do still matter in today's models.

An alternative line of work (option 2b) is \textit{auxiliary-task pre-training} and use labeled auxiliary tasks either via multi-task learning (MTL)~\cite{Peng2017} or intermediate-task transfer~\cite{stilts,phang2020english}. The latter proposed  \textit{supplementary training on intermediate labeled-data tasks for transfer} (STILT)~\cite{stilts}, and recently adopted this idea to cross-lingual learning, where  English is used as intermediate-task for zero-shot transfer~\cite{phang2020english}. 

The choice of data used for pre-training (or the auxiliary tasks) do matter. Current transformer models are trained on either large general data like BookCorpus and Wikipedia in BERT~\cite{devlin-etal-2019-bert} or target-specific samples, like papers from Semantic Scholar in SciBERT~\cite{beltagy-etal-2019-scibert}, and PubMed abstracts and PMC full-text articles in BioBERT~\cite{lee2020biobert}. What denotes \textit{relevant} data is an open question. Today, it is either general background knowledge, domain-specific target data, or a combination thereof, possibly via auxiliary tasks or intermediate training stages. Most of these have been carefully selected manually, raising interesting connections to data selection (Section~\ref{sec:dataselection}) and finding better curricula~\cite{tsvetkov-etal-2016-learning} to learn under domain shift~\cite{ruder-plank-2017-learning}.

While large pre-trained models have shown to work well, many questions and challenges remain. 
Recent work has shown that these models degrade on out-of-domain data, maximum likelihood training makes them too over-confident~\cite{oren-etal-2019-distributionally} and particularly calibration is important for out-of-domain generalization \cite{hendrycks2020pretrained}.  An acknowledged issue with fine-tuning is the brittleness of the process~\cite{stilts,dodge2020fine}. Even with the same hyperparameters, distinct runs can lead to drastically different results and training data order and seed choice have a considerably impact~\cite{dodge2020fine}. Deeper investigations into what such models capture, how they can be  robustly trained in light of known test distributions or out-of-domain conditions are interesting issues.  

\section{Hybrid approaches}\label{sec:hybrid}
Work on the intersection of data-centric and model-centric methods can be plentiful. It currently includes combining semi-supervised objectives with an adversarial loss~\cite{lim_aaai2020,alam-etal-2018-domain}, combining pivot-based approaches with pseudo-labeling~\cite{cui-bollegala-2019-self} and very recently with contextualized word embeddings~\cite{bendavid2020}, and combining multi-task approaches with domain shift~\cite{jia-etal-2019-cross}, multi-task learning with pseudo-labeling (multi-task tri-training)~\cite{ruder-plank-2018-strong}, and  \textit{adaptive ensembling}~\cite{desai-etal-2019-adaptive}, which uses a student-teacher network with a consistency-based self-ensembling loss and a temporal curriculum. They apply adaptive ensembling to study temporal and topic drift in  political data classification~\cite{desai-etal-2019-adaptive}.

\section{Challenges and future directions} \label{sec:challenges}

While recent work has made important progress in neural UDA, our survey reveals i) an over-representation and bias of work on sentiment analysis (cf.\ column bias in Table~\ref{tab:overview}) and ii) a general lack of testing across tasks (row sparsity in Table~\ref{tab:overview}) and multiple adaptation methods. 



\paragraph{Comprehensive UDA benchmarks} Concretely, we recommend a) to create new benchmarks for UDA with multiple tasks and of increasing complexity, setups beyond 1:1 adaptation, and datasets which document known \textit{variety facets} of the data~\cite{bender2018data}. This will help to learn about the known and unknown (Section~\ref{sec:varietyspace}) as `variety' (domain) matters; 
b) to release unlabeled data from the broader distribution from which annotated data was sampled, in line with~\newcite{gururangan2020dont}; this allows studying diachronic effects, as labeled evaluation data lacks diversity in terms of topics and time~\cite{desai-etal-2019-adaptive,derczynski-etal-2016-broad}; and c) to release unaggregated, multiple annotations to study divergences in annotations~\cite{plank-etal-2014-linguistically}.

\paragraph{Back to the roots and how knowledge transfers}  Revisiting classics in neural times is beneficial, as shown for example in recent work which brings back SCL and pseudo-labeling methods (see Table~\ref{tab:overview}), but much is left to see how these methods generalize. This can be linked to the question on what representations capture~\cite{probing-survey} and how knowledge transfers~\cite{rethmeier2019txray}. 

\paragraph{X scarcity}
Even unlabeled data can be scarce (X scarcity), particularly in highly-specialized language varieties (e.g., clinical data)~\cite{rethmeier-plank-2019-morty}. This is often due to data sharing restrictions. In some cases, only a trained source model could be available instead of raw or labeled texts~\cite{laparra2020rethinking}. Together with the quest for more efficient learning methods, the general question of how to adapt in light of X scarcity or absence becomes important.

\section{Conclusion} \label{sec:conclusions}
In this survey, we review  strands of unsupervised domain adaptation, summarized into \textit{model-centric}, \textit{data-centric}, and \textit{hybrid} methods, including trends in pre-training. We also revisit the notion of \textit{domain} and suggest to use the term \textit{variety} instead, to better capture the multitude of dimensions of variation. Our survey identifies a limited focus on sentiment benchmarks and single-task evaluation for UDA. Lastly, we outline future directions, linking to the broader challenges related to learning beyond 1:1 scenarios and out-of-distribution generalization. This also calls for new directions on benchmarks and learning under scarce data. 

\section*{Acknowledgements} 
We thank Nils Rethmeier, Raffaella Bernardi, Rob van der Goot and Sebastian Ruder for precious feedback on earlier drafts of this survey. This research is supported by a visit grant to Alan supported by COSBI and a research leader sapere aude grant to Barbara by the Independent Research Fund Denmark
 (Danmarks Frie Forskningsfond,  grant number 9063-00077B MultiVaLUe).  

\bibliographystyle{acl}
\bibliography{survey}

\newpage


\end{document}